\def\year{2020}\relax
\documentclass[letterpaper]{article} % DO NOT CHANGE THIS
\usepackage{aaai20}  % DO NOT CHANGE THIS
\usepackage{times}  % DO NOT CHANGE THIS
\usepackage{helvet} % DO NOT CHANGE THIS
\usepackage{courier}  % DO NOT CHANGE THIS
\usepackage[hyphens]{url}  % DO NOT CHANGE THIS
\usepackage{graphicx} % DO NOT CHANGE THIS
\usepackage{graphicx}  % DO NOT CHANGE THIS
\usepackage{multirow}
\usepackage{amsmath}
\usepackage{amssymb}
\usepackage{wasysym}
\usepackage{rotating}

\newcommand{\parencite}[1]{(\citeauthor{#1} \citeyear{#1})}
\providecommand{\tabularnewline}{\\}
\title{Universal-RCNN: Universal Object Detector via Transferable Graph R-CNN}
\author{Hang Xu,\textsuperscript{\rm 1 \thanks{Both authors contributed equally to this work.}} Linpu Fang,\textsuperscript{\rm 2 $^{*}$} Xiaodan Liang,\textsuperscript{\rm 3}\thanks{Corresponding Author: xdliang328@gmail.com}\\ \Large \textbf{Wenxiong Kang,\textsuperscript{\rm 2} Zhenguo Li\textsuperscript{\rm 1}} \\ 
	\textsuperscript{\rm 1}Huawei Noah's Ark Lab\\ 
	\textsuperscript{\rm 2}South China University of Technology\\ 
	\textsuperscript{\rm 3}Sun Yat-Sen University}

\begin{document}

\maketitle
\begin{abstract}
The dominant object detection approaches treat each dataset separately
and fit towards a specific domain, which cannot adapt to other domains
without extensive retraining. In this paper, we address the problem
of designing a universal object detection model that exploits diverse
category granularity from multiple domains and predict all kinds of
categories in one system. Existing works treat this problem by integrating
multiple detection branches upon one shared backbone network. However,
this paradigm overlooks the crucial semantic correlations between
multiple domains, such as categories hierarchy, visual similarity,
and linguistic relationship. To address these drawbacks, we present
a novel universal object detector called Universal-RCNN that incorporates
graph transfer learning for propagating relevant semantic information
across multiple datasets to reach semantic coherency. Specifically,
we first generate a global semantic pool by integrating all high-level
semantic representation of all the categories. Then an Intra-Domain
Reasoning Module learns and propagates the sparse graph representation
within one dataset guided by a spatial-aware GCN. Finally, an Inter-Domain
Transfer Module is proposed to exploit diverse transfer dependencies
across all domains and enhance the regional feature representation
by attending and transferring semantic contexts globally.  Extensive
experiments demonstrate that the proposed method significantly outperforms
multiple-branch models and achieves the state-of-the-art results on
multiple object detection benchmarks (mAP: 49.1\% on COCO). 
\end{abstract}

\section{Introduction}

\begin{figure}
\begin{centering}
\includegraphics[scale=0.33]{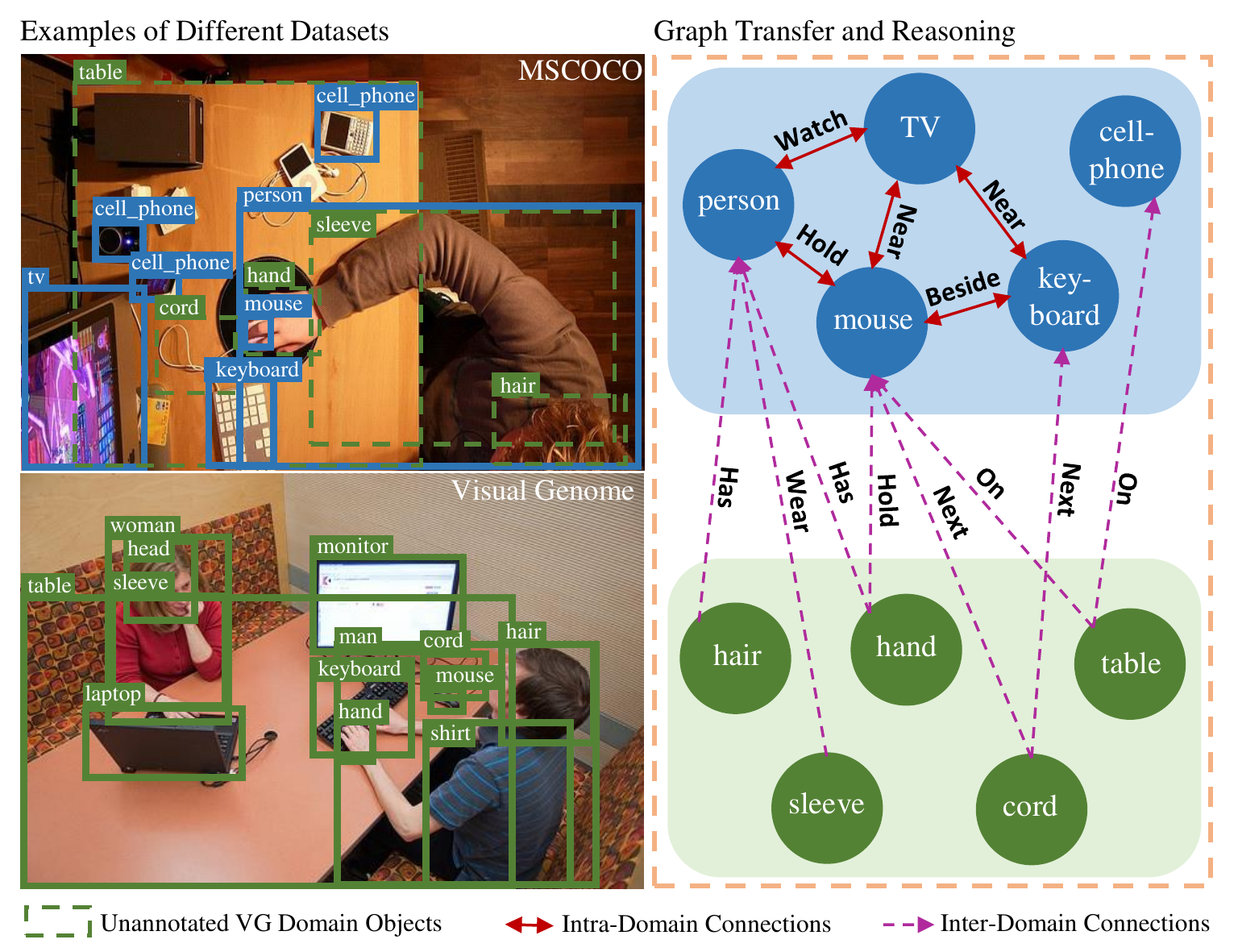}
\par\end{centering}
\caption{\label{fig:intro-graph}To identify objects in the upper image from
COCO, we could use graph transfer and reasoning across COCO and Visual
Genome (VG) (bottom): 1) Objects like ``TV'', and ``keyboard''
in the upper image could help to identify ``mouse'' and ``person'',
which requires reasoning and this inspires our Intra-Domain Reasoning
Module (blue). 2) Although ``table'', ``hand'', and ``cord''
are not annotated in the COCO image but present in VG, recognition
of them could help to recognize ``mouse'', ``keyboard'', which
motivates us the Inter-Domain Transfer Module (green to blue). }
\end{figure}

Object detection is a fundamental vision task that recognizes object's
location and category in an image. Modern object detectors widely
benefits autonomous vehicles, surveillance camera, mobile phone, to
name a few. According to the individual application, datasets annotated
with different categories were created to train a highly-specific
and distinct detector e.g. PASCAL VOC  \parencite{Everingham10} (20
categories), MSCOCO  \parencite{lin2014microsoft} (80 categories)
, Visual Genome (VG)  \parencite{krishnavisualgenome} (more than 33K
categories) and BDD  \parencite{yu2018bdd100k} (10 categories). Those
highly-tuned networks have sacrificed the generalization capability
and only fit towards each dataset domain. It is impossible to directly
adapt the model trained on one dataset to another related task, and
thus requires new data annotations and additional computation to train
each specific model. In contrast, human is capable of identifying
all kinds of objects precisely under complex circumstances and reaches
a holistic recognition across different domains. This may be due to
the remarkable reasoning and transferring ability about the relationship,
visual similarity and categories hierarchy across scenes. This inspires
us to explore how to endow the current detection system the ability
of incorporating and transferring relevant semantic information across
multiple domains in an effective way, in order to mimic human recognition
procedure. 

With different categories and granularity across domains, transferring
relevant information and reasoning among objects can help to make
a correct identification. An example of the necessity of transferring
relevant information between mutliple datasets and domains can be
found in Figure \ref{fig:intro-graph}.

The most widely-used and straightforward solutions to universal detection
would be to consider it as a multi-task learning problem, and integrate
multiple branches upon one shared backbone network  \parencite{he2017mask,gong2017look,li2016deepsaliency,liang2015human,dai2016instance}.
However, they overlook the crucial semantic correlations between multiple
domains and regions, such as categories hierarchy, visual similarity,
and linguistic relationship since feature-level information sharing
can only help to extract a more robust feature. Recently, some works
explore visual reasoning and try to combine different information
or interactions between objects in one image  \parencite{hu2017relation,chen2017spatial,wang2018non,liu2018structure}.
For example,  \citeauthor{jiang2018hybrid} recently try to incorporate semantic
relation reasoning in large-scale detection by different kinds of
knowledge forms.  \citeauthor{hu2017relation} introduced the Relation Networks
which use an adapted attention module to allow interaction between
the object's visual features. However, they did not consider inter-domain
relationship and their fully-connected relation is inefficient and
noisy by incorporating redundant and distracted relationships from
irrelevant objects and backgrounds. 

With the advancement of geometric deep learning  \parencite{bronstein2017geometric,monti2017geometric,velickovic2017graph},
using graph seems to be the most appropriate way to model relation
and interaction with its flexible structure. In this paper, we present
a novel universal detection system Universal-RCNN that incorporates
graph learning for propagating and transferring relevant semantic
information across multiple datasets to reach semantic coherency.
The proposed framework first generates a global semantic pool for
all domains to integrate all high-level semantic representation of
categories by distilling the weights of the object classifiers. Then
an Intra-Domain Reasoning Module learns a sparse regional graph to
encode the regional interaction and propagates the high-level semantic
graph representation from the global pool within one dataset guided
by a spatial-aware Graph Convolutional Neural Network (GCN). Furthermore,
an Inter-Domain Transfer Module is proposed to exploit diverse transfer
dependencies across all domains and enhance the regional feature representation
by attentively transferring related semantic contexts from the semantic
pool in one domain to another, which bridges the gap between domains,
and effectively utilize the annotations of multiple datasets. In this
work, we exploit various graph transfer dependencies such as attribute/relationship
knowledge, and visual/linguistic similarity. Our Universal-RCNN thus
enables adaptive global reasoning and transfers over regions in multiple
domains. The regional feature is greatly enhanced by abundant relevant
contextual Intra/Inter-domain information and the performance on each
domain is then boosted by sharing and distilling essential characteristics
across domains.

Extensive experiments demonstrate the effectiveness of the proposed
method and achieve the state-of-the-art results on multiple object
detection benchmarks. We observe a consistent gain over multiple-branch
models. In particular, Universal-RCNN achieves around 16\% of mAP
improvement on MSCOCO, 26\% on VG, and 36\% on ADE. The Universal-RCNN
obtains 49.1\% mAP on COCO \textit{test-dev} with single-model result.

\section{Related Work}

\textbf{Object Detection.} Object detection is a core problem in computer
vision. Most of the previous progress focus on developing new structures
such as better feature fusion  \parencite{lin2017feature,liu2018path,zhu2019feature,xu2019spatial}
and better receptive field to improve feature representation  \parencite{luo2017non,wang2018non,li2018detnet}.
However, their trained model cannot be applied directly to another
related task without heavy fine-tuning. 

\textbf{Transfer Learning.} Transfer learning  \parencite{zamir2018taskonomy,peng2018joint,cui2018large,xu2019reasoning}
tries to bridge the gap between different domains or tasks to reuse
the information and mitigate the burden of manual labeling. Early
work  \parencite{hoffman2014lsda} tries to learn to transfer from
the ImageNet's classification network into object detection network
with fewer categories. In  \citeauthor{misra2017red}, the
classifier weights are constructed by regression or a small neural
network from few-shot examples or different concepts. 

\textbf{Multi-task Learning.} Multi-task learning aims at developingRCNN
can sufficiently mining the semantic correlati systems that can provide
multiple outputs simultaneously for an input  \parencite{tsai2017learning,he2017mask,kirillov2018panoptic,xiong2019upsnet,kendall2018multi}.
For example, Mask-RCNN solves the instance segmentation problem by
considering two branches: one bounding box and classification head
and another dense image prediction head.  \citeauthor{xiao2018unified} introduced
a multi-task network to handle heterogeneous annotations for unified
perceptual scene parsing. However, these approaches simply create
several branches separately for different tasks. 

\section{The Proposed Approach}

\begin{figure*}
\begin{centering}
\includegraphics[scale=0.2]{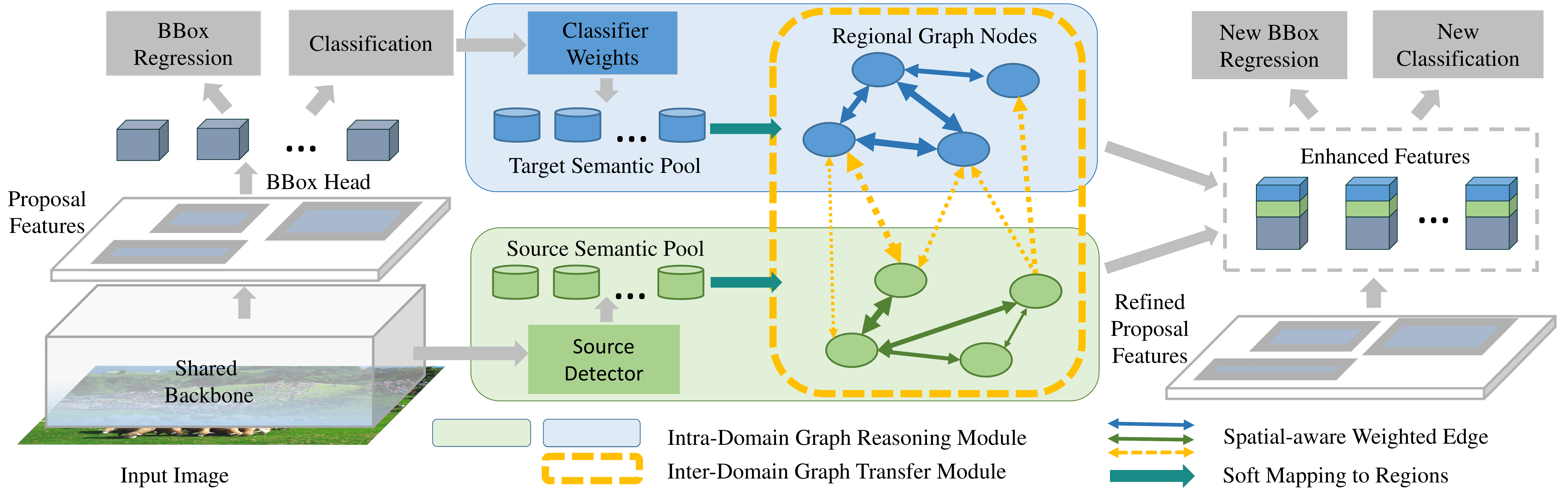}
\par\end{centering}
\caption{\label{fig:framework-graph}An overview of our Universal-RCNN. Built
on the classification layer of base detection network, global semantic
pools for each domain integrate all high-level semantic representation
for each category by the weights of previous classifiers. Then an
Intra-Domain Reasoning Module learns a sparse regional graph to encode
the regional interaction and propagates the high-level semantic graph
representation from the semantic pool within one domain. Furthermore,
Inter-Domain Transfer Module exploits diverse transfer dependencies
across domains and enhances the regional feature representation by
attentively transferring related semantic contexts from source semantic
pool to the target domain, which bridges the gap between domains and
utilizes the annotations of multiple datasets. Finally, the outputs
of modules concatenated with region proposal features are fed into
the bounding-box regression layer and classification layer to obtain
better detection results. Note that this figure only illustrates how
the model works within a specific pair of domains. In the whole model,
we have multiple pairs of the source/target domain for the inter-domain
transfer modules.}
\end{figure*}

\subsection{Overview}

In this paper, we introduce a unified detection framework to unify
all kinds of detection annotations from different domains and tackle
different detection tasks in one system, that is, detecting and predicting
all kinds of categories defined from multiple domains using a single
network while improving the detection performance of the individual
domain. This framework can be implemented on any modern dominant detection
system to further improve their performance by enhancing its original
image features via graph transfer learning. An overview of our model
can be found in Figure \ref{fig:framework-graph}. Specifically, we
first learn and propagate compact high-level semantic representation
within one domain via Intra-Domain Graph Reasoning module, and then
transfer and fuse the semantic information across multiple domains
to enhance the feature representation via Inter-Graph Transfer module
driven by learned transfer graph or human defined hierarchical categorical
structures.

\subsection{Intra-Domain Reasoning Module}

Given the extracted proposal visual features from the backbone, we
introduce Intra-Domain Reasoning module to enhance local features
for the domain \textbf{$\boldsymbol{T}$}, by leveraging graph reasoning
with key semantic and spatial relationships. More specifically, we
first create a global semantic pool to integrate high-level semantic
representation for each category by collecting the weights of the
original classification layer. Then a region-to-region undirected
graph $\boldsymbol{G}$ : $\boldsymbol{\boldsymbol{G}_{T\rightarrow T}}=<\mathcal{N},\mathcal{\mathcal{E}}>$
for domain \textbf{$\boldsymbol{T}$} is defined, where each node
in $\mathcal{N}$ corresponds to a region proposals and each edge
$e_{i,j}\in\mathcal{E}$ encodes relationship between two nodes. An
interpretable sparse adjacency matrix is learned from the visual feature
which only retains the most relevant connections for recognition of
the objects. Then semantic representations over the global semantic
pool are mapped to each region and propagated with a learned structure
of $\boldsymbol{\boldsymbol{G}_{T\rightarrow T}}$. Finally, the proposal
features are enhanced and concatenated with the original features
to obtain better detection results.

\subsubsection{\label{subsec:Learning-graph}Learning Graph $\boldsymbol{\boldsymbol{\boldsymbol{G}_{T\rightarrow T}}}$}

We seek to learn the $\mathcal{\mathcal{E}}_{T}\in\mathbf{\mathbb{R}}^{N_{r}\times N_{r}}$
in\textbf{ $\boldsymbol{\boldsymbol{\boldsymbol{G}_{T\rightarrow T}}}$}
thus the proposal node neighborhoods can be determined. We aim to
produce a graphical representation of the relationship (e.g. attribute
similarities and interactions) between proposal regions which is relevant
to the object detection. Given the regional visual features $\mathbf{f}=\{\boldsymbol{f}_{i}\}_{i=1}^{N_{r}},f_{i}\in\mathbf{\mathbb{R}}^{D}$
of $D$ dimension extracted from the backbone network, we first transform
$\mathbf{f}$ to a latent space $\mathbf{Z}$ by linear transformation
denoted by $z_{i}=\phi(\mathbf{f}),i=1,2,...,N_{r}$, where $z_{i}\in\mathbf{\mathbb{R}}^{L}$,
$L$ is the dimension of the latent space and $\phi(.)$ is a linear
function. Let $\mathbf{Z}\in\mathbf{\mathbb{R}}^{N_{r}\times L}$
be the collection of normalized $\{z_{i}\}_{i=1}^{N_{r}},z_{i}\in\mathbf{\mathbb{R}}^{L}$,
the adjacency matrix for $\boldsymbol{\boldsymbol{\boldsymbol{G}_{T\rightarrow T}}}$
with self loops can be calculated as $\mathcal{\mathcal{E}}=\mathbf{Z}\mathbf{Z}^{T}$,
so that $e_{i,j}=\frac{z_{i}z_{j}^{T}}{\left\Vert z_{i}z_{j}^{T}\right\Vert }$,
where $\left\Vert .\right\Vert $ is the L2-norm.

To determine the neighborhoods of each node, using a fully connected
$\mathcal{E}$ directly will establish relationship between backgrounds(negative)
samples which will lead to redundant edges and greater computation
cost. In this paper, we consider a sparse graph. For each region proposal
$i$, we only retain the top $t$ largest values of each row of $\mathcal{E}$,
that is: $\textrm{Neighbour}(\textrm{Node }i)=\textrm{Top-t}_{j=1,..,N_{r}}(e_{i,j})$.
This sparsity constraint ensures a spare graph structure focusing
on the most relevant relationship for recognition of the objects.

\subsubsection{Feature Enhanced via Intra-Domain Graph Reasoning. \label{subsec:Feature-Enhanced-via}}

Most recent existing works  \parencite{gong2018frage,chen2018iterative,jiang2018hybrid}
propagate visual features locally among regions in the image and the
information is only from those categories appearing in the image.
This limits the performance of graph reasoning because of poor or
distracted feature representations. Instead, we create a global semantic
pool\textbf{ }$\mathbf{P_{T}}$ to store high-level semantic representations
for all categories. In some works  \parencite{wang2018zero,gong2018frage,gidaris2018dynamic}
in zero/few-shot problem, they try to train a model to fit the weights
of the classifier of an unseen/unfamiliar category and the weights
of the classifier for each category can be regarded as containing
high-level semantic information. Formally, let $\mathbf{P_{T}\in\mathbf{\mathbb{R}}}^{C_{T}\times(D+1)}$
denote the weights and the bias of the previous classifier for all
the $C_{T}$ categories in the domain \textbf{$\boldsymbol{T}$}.
The global semantic pool\textbf{ }$\mathbf{P_{T}}$ is extracted from
the previous classification layer in the bbox head of the detection
network and can be updated in each iteration during training.

Since our graph $\boldsymbol{G}$ is a region-to-region graph, we
first need to map the category semantic embedding $p_{T}\in\mathbf{P_{T}}$
to the regional representations of nodes $x_{i}\in\boldsymbol{X}$
. In this paper, we use a \textit{soft-mapping} which computes the
mapping weights $m_{p_{T}\rightarrow x_{i}}\in\mathbf{M_{T}}$ as
$m_{p_{T}\rightarrow x_{i}}=\frac{\exp(s_{ij})}{\sum_{j}\exp(s_{ij})},$
where $s_{ij}$ is the classification score for the region $i$ towards
category $j$ from the previous classification layer of the detector.
Thus the regional representations of the nodes $\boldsymbol{X}\in\mathbf{\mathbb{R}}^{N_{r}\times(D+1)}$
can be computed as a matrix multiplication: $\boldsymbol{X}=\mathbf{M_{T}}\mathbf{P_{T}},$
where $\mathbf{M_{T}}\in\mathbf{\mathbb{R}}^{N_{r}\times C_{T}}$
is the soft-mapping matrix.

\begin{figure}[t]
\begin{centering}
\includegraphics[scale=0.28]{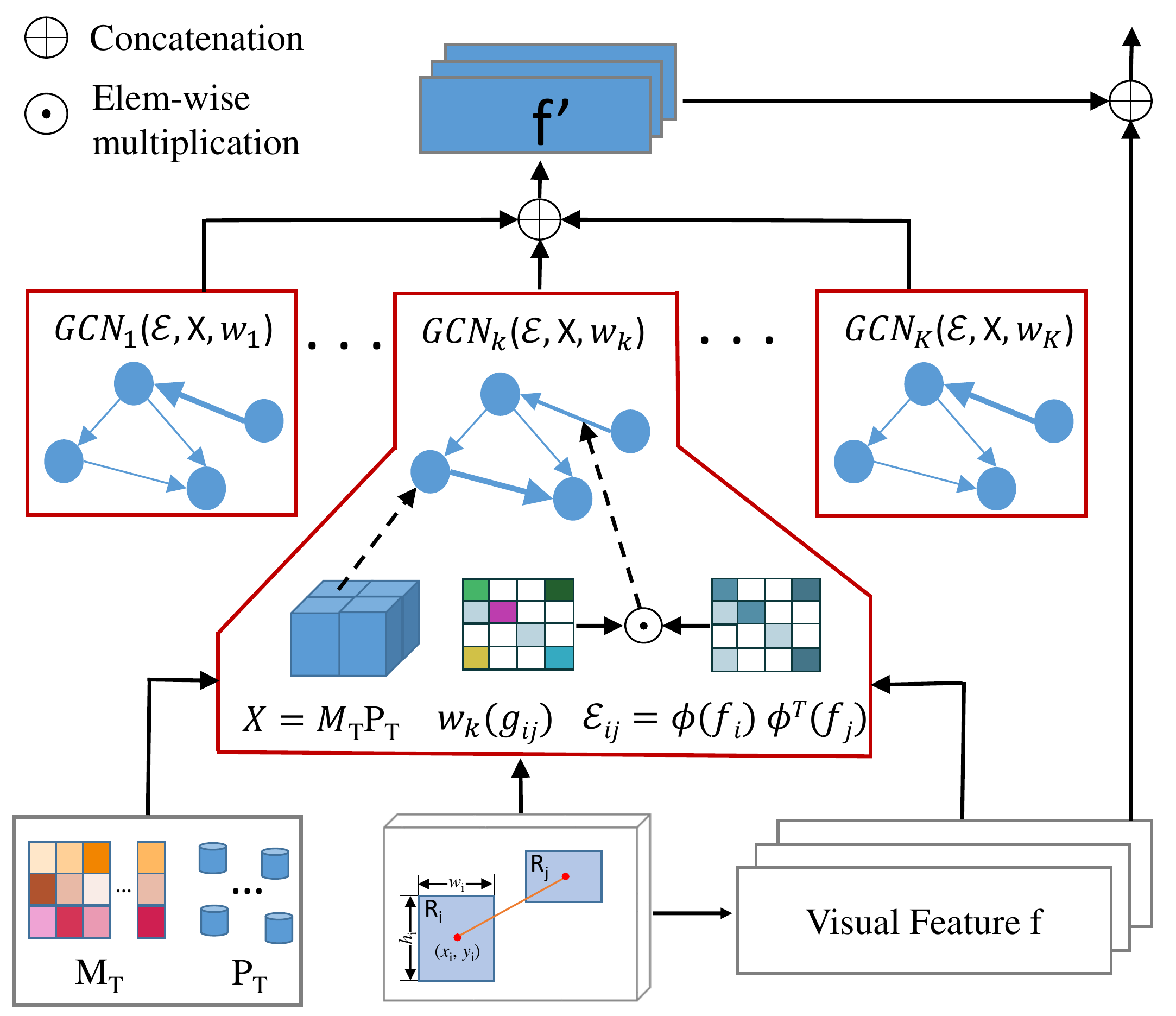}
\par\end{centering}
\caption{\label{fig:Flowchart}Flowchart of our graph reasoning and transfer
in Equation (1). The global semantic pool $\mathbf{P_{T}}$ is soft-mapped
to the regional nodes by $\boldsymbol{X}=\mathbf{M_{T}}\mathbf{P_{T}}$.
The graph edge $\mathcal{E}$ and spatial weight terms $w_{k}(g_{ij})$
are calculated from $\mathbf{f}$ and spatial information between
proposal $i$ and proposal $j$. Then graph conv $GCN_{k}(.)$ is
performed on the nodes $\boldsymbol{X}$ according to the graph edge
$\mathcal{E}$ and $w_{k}$. Final enhanced feature $\mathbf{f}'$
is a concatenation of all the GCN outputs. For inter-domain transfer
module, the idea is similar except the input graph edge: $\mathcal{\mathcal{E}}_{S\rightarrow T}$
and the nodes: $\boldsymbol{Y}=\mathbf{M}_{\mathbf{S}}\mathbf{P_{S}}$
which are transferred from source semantic pool.}
\end{figure}

Given the node representation $x_{j}\in\boldsymbol{X}$ and the learned
graph $\boldsymbol{\boldsymbol{\boldsymbol{G}_{T\rightarrow T}}}$,
it is natural to use a GCN for modeling the relation and interaction.
We thus define a patch operator for the GCN as follows:
\begin{equation}
\mathrm{f}_{k}'(i)=\sum_{j\in\textrm{Neighbour}(i)}w_{k}(g_{ij})x_{j}e_{ij},
\end{equation}
where $\textrm{Neighbour}(i)$ denotes the neighborhood of node $i$
and $e_{ij}$ is the normalized adjacency element of $\mathcal{\mathcal{E}}$.
To capture the pairwise spatial information between proposals, we
further add $K$ spatial weight terms $w_{k}(g_{ij})$ which are calculated
by a nonlinear transformation function $w_{k}(.)$ encoding the spatial
information of regions. $g_{ij}$ is a four dimensional relative geometry
feature between proposal $i$ and proposal $j$: $(\log\frac{\left|x_{i}-x_{j}\right|}{w_{i}},\log\frac{\left|x_{i}-x_{j}\right|}{h_{i}},\log\frac{w_{i}}{w_{j}},\log\frac{h_{i}}{h_{j}})$,
where $w_{i}$ and $h_{i}$ denotes the width and height of the region.
We consider $K$ set of $w_{k}(.)$ to encode different kinds of spatial
interactions. Flowchart of the graph reasoning can be found in Figure
\ref{fig:Flowchart}.

Then $\mathrm{f}'_{k}(i)$ for each node goes through a linear transformation
$\boldsymbol{L}\in\mathbf{\mathbb{R}}^{E\times(D+1)}$and is concatenated
together: $\mathbf{f}'=[L(\mathrm{f}_{k}'(i))]$, and $KE$ is the
dimension of the output enhanced feature for each region. Finally,
the $\mathbf{f}'$ for each region is concatenated to the original
region features $\mathbf{f}$ to improve both classification and localization.
Note that the $\mathbf{f}'$ is a distilled information across the
categories with connected edges such as similar attributes or relations.
Thus, sharing the common features between categories can help to improve
the feature representation by adding and discovering adaptive contexts
from the global semantic pool.

\subsection{Inter-Domain Transfer Module}

To effectively distill relevant information from datasets with different
annotations, we introduce Inter-Domain Transfer Module to bridge the
gap between different domains. This module can be applied for multiple
source domains as long as they have semantic correlation. More specifically,
for each pair of domains, we implement an inter-domain transfer module
from source domain $\boldsymbol{S}$ to target domain $\boldsymbol{T}$.
Given an image, the RPN head proposes a set of region proposals, some
of which may contain objects defined in other source domains that
have some relationships with objects defined in target domain $\boldsymbol{T}$.
We will first extract region proposals semantic features of a certain
source domain $\boldsymbol{S}$ according to the global semantic \textbf{$\mathbf{P_{S}}$}
pool of $\boldsymbol{S}$. Then these semantic features are transferred
to the target domain $\boldsymbol{T}$ by the GCN with the predefined
transfer graph $\boldsymbol{\boldsymbol{G}_{S\rightarrow T}}$ between
the domain $\boldsymbol{S}$ and $\boldsymbol{T}$. Finally, the proposal
features are concatenated by the information of the source domain
for better prediction.

\begin{table*}
\begin{centering}
\tabcolsep 0.01in{\scriptsize{}}%
\begin{tabular}{l|c|cccccc|cccccc}
\hline 
{\scriptsize{}\%} & {\scriptsize{}Method} & {\scriptsize{}AP} & {\scriptsize{}AP$_{50}$} & {\scriptsize{}AP$_{75}$} & {\scriptsize{}AP$_{S}$} & {\scriptsize{}AP$_{M}$} & {\scriptsize{}AP$_{L}$} & {\scriptsize{}AR$_{1}$} & {\scriptsize{}AR$_{10}$} & {\scriptsize{}AR$_{100}$} & {\scriptsize{}AR$_{S}$} & {\scriptsize{}AR$_{M}$} & {\scriptsize{}AR$_{L}$}\tabularnewline
\hline 
\multirow{3}{*}{{\scriptsize{}\begin{sideways}{\scriptsize{}MSCOCO}\end{sideways}}} & {\scriptsize{}FPN  \parencite{lin2017feature}} & {\scriptsize{}38.6} & {\scriptsize{}60.4} & {\scriptsize{}41.8} & {\scriptsize{}22.3} & {\scriptsize{}43.2} & {\scriptsize{}49.8} & {\scriptsize{}31.8} & {\scriptsize{}50.5} & {\scriptsize{}53.2} & {\scriptsize{}33.9} & {\scriptsize{}58.0} & {\scriptsize{}67.1}\tabularnewline
 & {\scriptsize{}Multi Branches} & {\scriptsize{}39.9} & {\scriptsize{}61.5} & {\scriptsize{}43.4} & {\scriptsize{}23.5} & {\scriptsize{}44.9} & {\scriptsize{}51.4} & {\scriptsize{}32.7} & {\scriptsize{}52.4} & {\scriptsize{}55.2} & {\scriptsize{}36.3} & {\scriptsize{}60.3} & {\scriptsize{}68.3}\tabularnewline
 & {\scriptsize{}Universal-RCNN} & \textbf{\scriptsize{}44.7}{\scriptsize{}$^{+6.1}$} & \textbf{\scriptsize{}65.1}{\scriptsize{}$^{+4.7}$} & \textbf{\scriptsize{}49.1}{\scriptsize{}$^{+7.3}$} & \textbf{\scriptsize{}26.6}{\scriptsize{}$^{+4.3}$} & \textbf{\scriptsize{}49.1}{\scriptsize{}$^{+5.9}$} & \textbf{\scriptsize{}58.6}{\scriptsize{}$^{+8.8}$} & \textbf{\scriptsize{}35.6}{\scriptsize{}$^{+3.8}$} & \textbf{\scriptsize{}56.6}{\scriptsize{}$^{+6.1}$} & \textbf{\scriptsize{}59.5}{\scriptsize{}$^{+6.3}$} & \textbf{\scriptsize{}39.4}{\scriptsize{}$^{+5.5}$} & \textbf{\scriptsize{}64.3}{\scriptsize{}$^{+6.3}$} & \textbf{\scriptsize{}75.1}{\scriptsize{}$^{+8.0}$}\tabularnewline
\hline 
\multirow{3}{*}{{\scriptsize{}\begin{sideways}{\scriptsize{}VG}\end{sideways}}} & {\scriptsize{}FPN  \parencite{lin2017feature}} & {\scriptsize{}7.0} & {\scriptsize{}12.3} & {\scriptsize{}7.0} & {\scriptsize{}3.9} & {\scriptsize{}7.6} & {\scriptsize{}10.2} & {\scriptsize{}14.0} & {\scriptsize{}19.1} & {\scriptsize{}19.3} & {\scriptsize{}10.5} & {\scriptsize{}19.4} & {\scriptsize{}22.5}\tabularnewline
 & {\scriptsize{}Multi Branches} & {\scriptsize{}7.1} & {\scriptsize{}12.3} & {\scriptsize{}7.2} & {\scriptsize{}4.1} & {\scriptsize{}7.6} & {\scriptsize{}10.3} & {\scriptsize{}13.8} & {\scriptsize{}19.0} & {\scriptsize{}19.2} & {\scriptsize{}10.9} & {\scriptsize{}19.0} & {\scriptsize{}22.2}\tabularnewline
 & {\scriptsize{}Universal-RCNN} & \textbf{\scriptsize{}8.8}{\scriptsize{}$^{+1.8}$} & \textbf{\scriptsize{}14.2}{\scriptsize{}$^{+1.9}$} & \textbf{\scriptsize{}9.3}{\scriptsize{}$^{+2.3}$} & \textbf{\scriptsize{}5.0}{\scriptsize{}$^{+1.1}$} & \textbf{\scriptsize{}9.4}{\scriptsize{}$^{+1.8}$} & \textbf{\scriptsize{}13.3}{\scriptsize{}$^{+3.1}$} & \textbf{\scriptsize{}17.5}{\scriptsize{}$^{+3.5}$} & \textbf{\scriptsize{}23.7}{\scriptsize{}$^{+4.6}$} & \textbf{\scriptsize{}24.0}{\scriptsize{}$^{+4.7}$} & \textbf{\scriptsize{}13.4}{\scriptsize{}$^{+2.9}$} & \textbf{\scriptsize{}23.3}{\scriptsize{}$^{+3.9}$} & \textbf{\scriptsize{}28.8}{\scriptsize{}$^{+6.3}$}\tabularnewline
\hline 
\multirow{3}{*}{{\scriptsize{}\begin{sideways}{\scriptsize{}ADE}\end{sideways}}} & {\scriptsize{}FPN  \parencite{lin2017feature}} & {\scriptsize{}11.3} & {\scriptsize{}19.6} & {\scriptsize{}11.7} & {\scriptsize{}6.9} & {\scriptsize{}11.8} & {\scriptsize{}17.6} & {\scriptsize{}12.8} & {\scriptsize{}19.5} & {\scriptsize{}20.1} & {\scriptsize{}12.6} & {\scriptsize{}21.4} & {\scriptsize{}28.1}\tabularnewline
 & {\scriptsize{}Multi Branches} & {\scriptsize{}12.2} & {\scriptsize{}19.9} & {\scriptsize{}12.8} & {\scriptsize{}7.2} & {\scriptsize{}12.0} & {\scriptsize{}19.6} & {\scriptsize{}13.3} & {\scriptsize{}20.2} & {\scriptsize{}20.8} & {\scriptsize{}13.3} & {\scriptsize{}22.2} & {\scriptsize{}29.4}\tabularnewline
 & {\scriptsize{}Universal-RCNN} & \textbf{\scriptsize{}15.4}{\scriptsize{}$^{+4.1}$} & \textbf{\scriptsize{}24.2}{\scriptsize{}$^{+4.6}$} & \textbf{\scriptsize{}16.7}{\scriptsize{}$^{+5.0}$} & \textbf{\scriptsize{}9.4}{\scriptsize{}$^{+2.5}$} & \textbf{\scriptsize{}15.5}{\scriptsize{}$^{+3.7}$} & \textbf{\scriptsize{}24.3}{\scriptsize{}$^{+6.7}$} & \textbf{\scriptsize{}17.4}{\scriptsize{}$^{+4.6}$} & \textbf{\scriptsize{}26.1}{\scriptsize{}$^{+6.6}$} & \textbf{\scriptsize{}26.7}{\scriptsize{}$^{+6.6}$} & \textbf{\scriptsize{}17.3}{\scriptsize{}$^{+4.7}$} & \textbf{\scriptsize{}28.3}{\scriptsize{}$^{+6.9}$} & \textbf{\scriptsize{}37.1}{\scriptsize{}$^{+9.0}$}\tabularnewline
\hline 
\end{tabular}{\scriptsize\par}
\par\end{centering}
\caption{\label{tab:Main-results-onVG_ADE}Main results on MSCOCO(\textit{minival}),
VG, and ADE. ``Universal-RCNN'' is our full model trained on all
the three domains via graph transfer and reasoning. The backbones
of all the models are ResNet-101.}
\end{table*}

\subsubsection{Construct $\boldsymbol{\boldsymbol{G}_{S\rightarrow T}}$}

For different detection task with diverse categories, the attribute
similarities, relationships or hierarchical correlations among them
can be exploited. For example, person label in a COCO dataset contains
head, arm, and leg in VG domain, and the person label can also be
composed of more fine-grained categories (e.g., man, woman, boy and
girl) in ADE dataset. Also, there may further exists some location
and co-occurrence relationship such as ``road \& car'', ``street
\& truck'' and ``handbag \& arm''. Thus, to transfer information
from source domain to target domain, we define a transfer graph $\boldsymbol{\boldsymbol{G}_{S\rightarrow T}}$,
where $t_{c_{i}c_{j}}\in$$\boldsymbol{\boldsymbol{G}_{S\rightarrow T}}$
and $c_{i}$, $c_{j}$ denote the categories in source domain and
target domain respectively.

In this paper we further try different methods to construct graph
to explore various graph transfer dependencies between different label
sets. Thus we consider and compare four schemes for $\boldsymbol{\boldsymbol{G}_{S\rightarrow T}}$
including learning the graph from feature, handcrafted attribute/relationship,
and word embedding similarity. 

\textbf{Handcrafted Attribute.} Following  \citeauthor{jiang2018hybrid},
we consider a similar way to construct a handcrafted attribute $\boldsymbol{\boldsymbol{G}_{S\rightarrow T}}$.
Let us consider $K$ attributes such as colors, size, materials, and
status, we obtain a $(C_{S}+C_{T})\times K$ frequency distribution
table for each class-attribute pair. Then the pairwise Jensen\textendash Shannon
(JS) divergence between probability distributions $P_{c_{i}}$ and
$P_{c_{j}}$ of two categories $i$ and $j$ from source/target domain
can be measured as the edge weights of two classes: $t_{c_{i}c_{j}}=JS(P_{c_{i}}||P_{c_{j}})$,
where JS divergence measures the similarity between two distributions.

\textbf{Handcrafted Relationship.} We may consider the pairwise relationship
between classes, such as location relationship (e.g. \emph{along},
\emph{on}), the ``subject-verb-object'' relationship (e.g. \emph{eat},
\emph{wear}) or co-occurrence relationship. We can calculate frequent
statistics $f_{ij}$ from the occurrence among all source-target categories
pairs from additional linguistic information or through counting from
images. The symmetric transformation and row normalization are performed
on edge weights: $t_{c_{i}c_{j}}=\frac{f_{ij}}{\sqrt{d_{ii}d_{jj}}}$,
where $d_{ii}=\sum_{j}f_{ij}$.

\textbf{Word Embedding Similarity.} We further explore the linguistic
knowledge to construct the $\boldsymbol{\boldsymbol{G}_{S\rightarrow T}}$
besides the visual information. We use the word2vec model  \citeauthor{almazan2014word}
to map the semantic word of categories to a word embedding vector.
Then we compute the similarity between the names of the categories
in source domain and target domain: $t_{c_{i}c_{j}}=\frac{\exp(cos(w_{i},w_{j}))}{\sum_{j}\exp(cos(w_{i},w_{j}))}$,
where $cos(w_{i},w_{j})$ is the cosine similarity between the word
embedding vectors of the $i$th source category and $j$th target
category.

\textbf{Learning the Graph from Features. }This scheme is almost the
same as the one used in the Intra-Domain Reasoning Module. The visual
feature is transformed to the latent space $z$. Edge $\mathcal{\mathcal{E}}_{S\rightarrow T}$
between the source and target domain can be calculated by $\frac{z_{i}z_{j}^{T}}{\left\Vert z_{i}z_{j}^{T}\right\Vert }$.

\subsubsection{Feature Enhanced via Inter-Domain Transfer}

After creating a global semantic pool $\mathbf{P_{S}}$ for all the
$C_{S}$ categories, it is then natural to transfer high level semantic
information from the source domain $\mathbf{P_{S}\in\mathbf{\mathbb{R}}}^{C_{S}\times(D+1)}$
to the target domain by the transfer graph $\boldsymbol{\boldsymbol{G}_{S\rightarrow T}}$$\mathbf{\in\mathbf{\mathbb{R}}}^{C_{S}\times C_{T}}$.
Since the nodes are proposal regions, a $N_{r}\times N_{r}$ adjacent
matrix $\mathcal{\mathcal{E}}_{S\rightarrow T}$ between regions can
be obtained by a matrix multiplication: $\mathcal{\mathcal{E}}_{S\rightarrow T}=\mathbf{M}_{\mathbf{S}}\boldsymbol{G}_{S\rightarrow T}\mathbf{M_{T}}^{T}$
where $\mathbf{M}_{\mathbf{S}}\mathbf{\in\mathbf{\mathbb{R}}}^{N_{r}\times C_{S}}$
which is created similar to the soft-mapping by the classification
score for the region towards the source categories in Section \ref{subsec:Feature-Enhanced-via}.
Then we can also obtain regional representations of nodes $y_{i}\in\boldsymbol{Y}$
from the relevant information from source domain by $\boldsymbol{Y}=\mathbf{M}_{\mathbf{S}}\mathbf{P_{S}}$.

Finally, given the node representation $y_{i}\in\boldsymbol{Y}$ and
the adjacent matrix $\mathcal{\mathcal{E}}_{S\rightarrow T}\mathbf{\in\mathbf{\mathbb{R}}^{N_{r}\times N_{r}}}$,
we can also use weighted GCN as \ref{subsec:Feature-Enhanced-via}
to propagate the semantic representations from the source domain to
the target region nodes. Finally, the output of the GCN $\mathbf{f}_{s}'$
is concatenated to the original visual features $\mathbf{f}$ for
better classification and bounding box regression. Note that $\mathbf{f}_{s}'$
serves as supplementary information since the relevant prediction
in the source domain is transferred to each region to help better
recognize the item.

\section{Experiments}

\begin{table}
\begin{centering}
\tabcolsep 0.01in{\scriptsize{}}%
\begin{tabular}{c|cc|c|c|cc|c}
\hline 
{\scriptsize{}Eval} & {\scriptsize{}Methods} & {\scriptsize{}Train with} & {\scriptsize{}mAP} & {\scriptsize{}Eval} & {\scriptsize{}Methods} & {\scriptsize{}Train with} & {\scriptsize{}mAP}\tabularnewline
\hline 
\multirow{9}{*}{{\footnotesize{}\begin{sideways}{\scriptsize{}MSCOCO}\end{sideways}}} & {\scriptsize{}FPN} & {\scriptsize{}-} & {\scriptsize{}38.6} & \multirow{8}{*}{{\footnotesize{}\begin{sideways}{\scriptsize{}VG}\end{sideways}}} & {\scriptsize{}FPN} & {\scriptsize{}-} & {\scriptsize{}7.0}\tabularnewline
 & {\scriptsize{}Multi-Branches} & {\scriptsize{}VG} & {\scriptsize{}39.8} &  & {\scriptsize{}Multi-Branches} & {\scriptsize{}COCO} & {\scriptsize{}7.0}\tabularnewline
 & {\scriptsize{}Fine-tuning} & {\scriptsize{}VG} & {\scriptsize{}39.2} &  & {\scriptsize{}Fine-tuning} & {\scriptsize{}COCO} & {\scriptsize{}7.4}\tabularnewline
 & {\scriptsize{}Overlap Labels} & {\scriptsize{}VG} & {\scriptsize{}38.7} &  & {\scriptsize{}Universal-RCNN} & {\scriptsize{}COCO} & \textbf{\scriptsize{}8.2}\tabularnewline
\cline{6-8} \cline{7-8} \cline{8-8} 
 & {\scriptsize{}Pseudo Labels} & {\scriptsize{}VG} & {\scriptsize{}38.7} &  & {\scriptsize{}FPN} & {\scriptsize{}-} & {\scriptsize{}7.0}\tabularnewline
 & {\scriptsize{}Universal-RCNN} & {\scriptsize{}VG} & \textbf{\scriptsize{}43.5} &  & {\scriptsize{}Multi-Branches} & {\scriptsize{}ADE} & {\scriptsize{}7.0}\tabularnewline
\cline{2-4} \cline{3-4} \cline{4-4} 
 & {\scriptsize{}Multi-Branches} & {\scriptsize{}ADE} & {\scriptsize{}38.8} &  & {\scriptsize{}Fine-tuning} & {\scriptsize{}ADE} & {\scriptsize{}7.3}\tabularnewline
 & {\scriptsize{}Fine-tuning} & {\scriptsize{}ADE} & {\scriptsize{}38.6} &  & {\scriptsize{}Universal-RCNN} & {\scriptsize{}ADE} & \textbf{\scriptsize{}8.0}\tabularnewline
\cline{5-8} \cline{6-8} \cline{7-8} \cline{8-8} 
 & {\scriptsize{}Universal-RCNN} & {\scriptsize{}ADE} & \textbf{\scriptsize{}41.5} & \multirow{4}{*}{{\footnotesize{}\begin{sideways}{\scriptsize{}ADE}\end{sideways}}} & {\scriptsize{}FPN} & {\scriptsize{}-} & {\scriptsize{}11.3}\tabularnewline
\cline{1-4} \cline{2-4} \cline{3-4} \cline{4-4} 
\multirow{3}{*}{{\footnotesize{}\begin{sideways}{\scriptsize{}ADE}\end{sideways}}} & {\scriptsize{}Multi-Branches} & {\scriptsize{}VG} & {\scriptsize{}12.3} &  & {\scriptsize{}Multi-Branches} & {\scriptsize{}COCO} & {\scriptsize{}11.4}\tabularnewline
 & {\scriptsize{}Fine-tuning} & {\scriptsize{}VG} & {\scriptsize{}12.8} &  & {\scriptsize{}Fine-tuning} & {\scriptsize{}COCO} & {\scriptsize{}11.9}\tabularnewline
 & {\scriptsize{}Universal-RCNN} & {\scriptsize{}VG} & \textbf{\scriptsize{}14.6} &  & {\scriptsize{}Universal-RCNN} & {\scriptsize{}COCO} & \textbf{\scriptsize{}12.9}\tabularnewline
\hline 
\end{tabular}{\scriptsize\par}
\par\end{centering}
\caption{\label{tab:two-domain}Results of mAP with models trained based on
two domains. ``Universal-RCNN'' is our full model trained on two
domains via graph transfer and reasoning. The backbones are ResNet-101.}
\end{table}

\begin{table}[h]
\begin{centering}
\tabcolsep 0.01in{\scriptsize{}}%
\begin{tabular}{c|cccc|ccc|ccc}
\hline 
{\scriptsize{}Intra-Domain} & \multicolumn{4}{c|}{{\scriptsize{}Inter-Domain Trasfer}} & \multirow{2}{*}{{\scriptsize{}AP}} & \multirow{2}{*}{{\scriptsize{}AP$_{50}$}} & \multirow{2}{*}{{\scriptsize{}AP$_{75}$}} & \multirow{2}{*}{{\scriptsize{}AP$_{S}$}} & \multirow{2}{*}{{\scriptsize{}AP$_{M}$}} & \multirow{2}{*}{{\scriptsize{}AP$_{L}$}}\tabularnewline
\cline{2-5} \cline{3-5} \cline{4-5} \cline{5-5} 
{\scriptsize{}Reasoning} & {\scriptsize{}Attribute} & {\scriptsize{}Relation} & {\scriptsize{}Embed} & {\scriptsize{}Learn} &  &  &  &  &  & \tabularnewline
\hline 
 &  &  &  &  & {\scriptsize{}38.6} & {\scriptsize{}60.4} & {\scriptsize{}41.8} & {\scriptsize{}22.3} & {\scriptsize{}43.2} & {\scriptsize{}49.8}\tabularnewline
\textbf{\small{}$\checked$} &  &  &  &  & {\scriptsize{}41.4} & {\scriptsize{}60.9} & {\scriptsize{}46.0} & {\scriptsize{}22.9} & {\scriptsize{}46.1} & {\scriptsize{}55.2}\tabularnewline
\textbf{\small{}$\checked$} & \textbf{\small{}$\checked$} &  &  &  & {\scriptsize{}43.0} & {\scriptsize{}62.8} & {\scriptsize{}47.3} & {\scriptsize{}25.0} & {\scriptsize{}47.6} & {\scriptsize{}56.3}\tabularnewline
\textbf{\small{}$\checked$} &  & \textbf{\small{}$\checked$} &  &  & {\scriptsize{}43.2} & {\scriptsize{}63.0} & {\scriptsize{}47.3} & {\scriptsize{}25.4} & {\scriptsize{}47.5} & {\scriptsize{}56.8}\tabularnewline
\textbf{\small{}$\checked$} &  &  & \textbf{\small{}$\checked$} &  & {\scriptsize{}42.7} & {\scriptsize{}62.7} & {\scriptsize{}47.4} & {\scriptsize{}24.9} & {\scriptsize{}47.4} & {\scriptsize{}56.3}\tabularnewline
 &  &  &  & \textbf{\small{}$\checked$} & {\scriptsize{}41.9} & {\scriptsize{}60.9} & {\scriptsize{}46.1} & {\scriptsize{}23.5} & {\scriptsize{}46.2} & {\scriptsize{}55.8}\tabularnewline
\textbf{\small{}$\checked$} &  &  &  & \textbf{\small{}$\checked$} & \textbf{\scriptsize{}43.5} & \textbf{\scriptsize{}63.5} & \textbf{\scriptsize{}47.7} & \textbf{\scriptsize{}25.8} & \textbf{\scriptsize{}47.8} & \textbf{\scriptsize{}57.0}\tabularnewline
\hline 
\end{tabular}{\scriptsize\par}
\par\end{centering}
\caption{\label{tab:Ablation-study-on}Ablation study on MSCOCO(\textit{minival})
with models trained based on VG and MSCOCO domains. We consider four
schemes to construct the transfer dependencies including handcrafted
attribute(``Attribute''), relationship(``Relation''), word embedding
similarity(``Embed'') and learning the graph from features(``Learn'').
The backbones of all the models are ResNet-101.}
\end{table}

\textbf{Datasets and Evaluations.} We evaluate the performance of
our Universal-RCNN on three object detection domains with different
annotations of categories: MSCOCO 2017  \parencite{lin2014microsoft},
Visual Genome(VG)  \parencite{krishnavisualgenome}, and ADE  \parencite{zhou2017scene}.
MSCOCO is a common object detection dataset with 80 object classes,
which contains 118K training images, 5K validation images (denoted
as \textit{minival}) and 20K unannotated testing images (denoted as
\textit{test-dev}) as common practice. VG and ADE are two large-scale
object detection benchmarks with thousands of object classes. For
VG, we use the synsets  \parencite{russakovsky2015imagenet} instead
of the raw names of the categories due to inconsistent label annotations.
We use 88K images for training and 5K images for testing with 1000
most frequent classes, following  \citeauthor{chen2018iterative,jiang2018hybrid}.
For ADE, we consider 445 classes and use 20K images for training and
1K images for testing, following  \citeauthor{chen2018iterative,jiang2018hybrid}.
Since ADE is a segmentation dataset, we convert segmentation masks
to bounding boxes for all instances. For all the evaluation, we use
the standard COCO metrics including mean Average Precision (mAP) and
Average Recall (AR).

\textbf{Implementation Details. }The proposed Universal-RCNN is a
single network and trained in an end-to-end style. All tasks for multiple
domains share a backbone for intermediate image feature extraction
while having separated heads and transfer modules for multi-domain
object feature learning. We use the popular FPN as our baseline detector
and implement Universal-RCNN based on it. ResNet-101  \parencite{he2016deep}
pretrained on ImageNet is used as the shared backbone network. The
hyper-parameters in training mostly follow  \citeauthor{lin2017feature}.
During both training and testing, we resize the input image such that
the shorter side has 800 pixels. RPN and bbox-head are applied to
all levels in the feature pyramid. The total number of proposed regions
after NMS is $N_{r}=512$. In the bbox-head, 2 shared FC layer is
used for proposal visual feature extraction and the output is a 1024-d
vector feed into the bbox regression and class-agnostic classification
layer following FPN.

For the Intra-Domain graph reasoning and Inter-Domain transfer module,
we use the previous 2 shared FC layer for re-extracting visual features
$\mathbf{f}$ of region proposals. In the graph learner module, we
use a linear transformation layer of size $256$ to learns the latent
representation $\mathbf{Z}$ ($L=256$) and most $t=32$ relevant
nodes are retained. The global semantic pools are created for each
domain by copying the weights of the classification layers of different
domains. For the spatial-aware GCN, we use two weighted graph convolutional
layers with dimensions of $256$ and $128$ respectively so that the
output size of the module for each region is $128$. Each GCN consists
of $K=8$ spatial weight terms forming a multi-head graph attention
layer  \parencite{monti2017geometric}. All experiments are conducted
on a single server with 8 Tesla V100 GPUs by using the Pytorch framework.
For training, SGD with weight decay of 0.0001 and momentum of 0.9
is adopted to optimize all models. The batch size is set to be 16
with 2 images on each GPU. The initial learning rate is 0.02, reduce
twice (x0.1) during the training process. We train 12 epochs for all
models in an end-to-end manner. We follow  \citeauthor{jiang2018hybrid}
to prepare handcrafted Attribute/Relationship matrix with the help
of the annotations in the VG dataset.

\begin{table}
\begin{centering}
\tabcolsep 0.01in{\scriptsize{}}%
\begin{tabular}{c|c|ccc|ccc}
\hline 
{\scriptsize{}\# training} & {\scriptsize{}Method} & {\scriptsize{}AP} & {\scriptsize{}AP$_{50}$} & {\scriptsize{}AP$_{75}$} & {\scriptsize{}AP$_{S}$} & {\scriptsize{}AP$_{M}$} & {\scriptsize{}AP$_{L}$}\tabularnewline
\hline 
{\scriptsize{}50\%} & {\scriptsize{}FPN w Finetune} & {\scriptsize{}36.3} & {\scriptsize{}57.3} & {\scriptsize{}39.3} & {\scriptsize{}20.6} & {\scriptsize{}40.3} & {\scriptsize{}47.4}\tabularnewline
{\scriptsize{}MSCOCO} & {\scriptsize{}Multi Branches} & {\scriptsize{}37.6} & {\scriptsize{}59.0} & {\scriptsize{}40.8} & {\scriptsize{}21.8} & {\scriptsize{}42.6} & {\scriptsize{}48.4}\tabularnewline
{\scriptsize{}Data} & {\scriptsize{}Universal-RCNN} & \textbf{\scriptsize{}42.0} & \textbf{\scriptsize{}62.1} & \textbf{\scriptsize{}46.3} & \textbf{\scriptsize{}24.4} & \textbf{\scriptsize{}46.3} & \textbf{\scriptsize{}56.4}\tabularnewline
\hline 
{\scriptsize{}80\%} & {\scriptsize{}FPN w Finetune} & {\scriptsize{}38.3} & {\scriptsize{}59.4} & {\scriptsize{}41.8} & {\scriptsize{}21.8} & {\scriptsize{}42.6} & {\scriptsize{}50.0}\tabularnewline
{\scriptsize{}MSCOCO} & {\scriptsize{}Multi Branches} & {\scriptsize{}39.2} & {\scriptsize{}60.5} & {\scriptsize{}42.7} & {\scriptsize{}23.5} & {\scriptsize{}44.0} & {\scriptsize{}50.7}\tabularnewline
{\scriptsize{}Data} & {\scriptsize{}Universal-RCNN} & \textbf{\scriptsize{}42.9} & \textbf{\scriptsize{}63.1} & \textbf{\scriptsize{}46.8} & \textbf{\scriptsize{}25.1} & \textbf{\scriptsize{}47.0} & \textbf{\scriptsize{}56.5}\tabularnewline
\hline 
{\scriptsize{}100\%} & {\scriptsize{}FPN w Finetune} & {\scriptsize{}39.2} & {\scriptsize{}60.5} & {\scriptsize{}42.5} & {\scriptsize{}22.8} & {\scriptsize{}43.8} & {\scriptsize{}50.7}\tabularnewline
{\scriptsize{}MSCOCO} & {\scriptsize{}Multi Branches} & {\scriptsize{}39.8} & {\scriptsize{}61.2} & {\scriptsize{}43.2} & {\scriptsize{}23.1} & {\scriptsize{}44.3} & {\scriptsize{}51.1}\tabularnewline
{\scriptsize{}Data} & {\scriptsize{}Universal-RCNN} & \textbf{\scriptsize{}43.5} & \textbf{\scriptsize{}63.5} & \textbf{\scriptsize{}47.7} & \textbf{\scriptsize{}25.8} & \textbf{\scriptsize{}47.8} & \textbf{\scriptsize{}57.0}\tabularnewline
\hline 
\end{tabular}{\scriptsize\par}
\par\end{centering}
\caption{\label{tab:Training-with-less}Training with less data on MSCOCO(\textit{minival})
with models trained based on VG and MSCOCO domains. We only use a
portion of MSCOCO data to train the models. ``FPN w Finetune'' is
the FPN model first trained on VG, then fintuned on MSCOCO. ``FPN
w Multi Branches'' is the method FPN with two branches.}
\end{table}

\textbf{Comparison with state-of-the-art domain transfer baselines.
}To show the effecitveness of our method in transfering relevant information
across multiple domains, we compare the overall performance of Universal-RCNN
with several state-of-the-art domain transfer baselines including
a) training an unified model on all datasets with separate RPN and
bbox heads for each domain stacked on a shared backbone (\textbf{Multiple
Branches}); b) pretraining on source dataset, fine-tuning on target
dataset (\textbf{Fine-tuning}); c) training based on the overlap labels
(\textbf{Overlap Labels}); d) using a fully trained model from target
dataset and generate pseudo-labels on source images and train again
(\textbf{Pseudo-labels}). The comparison results are reported on MSCOCO,
VG and ADE in Table \ref{tab:Main-results-onVG_ADE} and Table \ref{tab:two-domain}. 

Table \ref{tab:Main-results-onVG_ADE} shows the results of our Universal
R-CNN that exploits the semantic correlations between all three datasets.
We only compare our Universal R-CNN with Multiple Branches method
under this setting since b), c), d) can only be used on two datasets.
As can be seen, the Universal-RCNN significantly outperforms Multiple
Branches method in terms of precision (1.7\% to 4.8\%) and recall
(4.7\% to 6.6\%), which indicates the superiority of explicitely extracting
the semantic correlatons across domains within our Universal-RCNN
than implicitly learning feature-level interactions between domains
within multi-task learning. 

Table \ref{tab:two-domain} shows the experiment results of training
on each pair of domains. It can be found that our method consistently
surpasses the above-mentioned domain transfer methods by a large margin
through graph reasoning and transfer. Both Multiple Branches method
and Fine-tuning method cannot improve detection performance much,
which demonstrates the implicit way to utilize semantic information
across multiple domains is ineffective. Since VG domain nearly contains
all class annotations of COCO domain (79 classes out of total 80 classes),
we test Overlap Labels method and Pseudo Labels method on COCO, the
results show that both of them only obtain negligible 0.1\% AP improvement.
The reason is that the classes of VG domain have many fine-grained
objects while the classes of COCO domain only have some specific objects.
Table \ref{tab:two-domain} also shows that transferring information
from VG to COCO or ADE can achieve better detection performance, this
is because that VG annotations contains a huge number of classes and
many of them overlap with that of COCO and ADE. Overall, our Universal-RCNN
can sufficiently mining the semantic correlations across all three
domains and significantly boost the detection performance on each
target domain.

\textbf{Ablative Analysis. }Table \ref{tab:Ablation-study-on} shows
the ablative analysis of our method trained based on VG and MSCOCO
domains. Our Intra-domain reasoning is effective and acquires 2.8\%
improvements compared with the basic network on MSCOCO. For the Inter-domain
transfer module, it can be found that adding the transfer module alone
can boost the performance by 3.3\% which demonstrates the importance
of graph transfer across domains. Combining these two modules can
further improve the performance by 1.6\%, which demonstrates the effectiveness
of the proposed Universal-RCNN in fully exploiting semantic correlations
across multiple domains. We further compare four schemes to construct
the transfer dependencies including handcrafted attribute, relationship,
word embedding similarity and learning the graph from feature. Learning
the graph from features performs better than other methods. Thus,
we choose learning the graph from feature for the final model.

\textbf{Training with Less Data.} Since bounding box annotation is
expensive, we are curious about how our Universal-RCNN performs with
less training data. Table \ref{tab:Training-with-less} shows the
performance on MSCOCO with models trained based on VG and MSCOCO(less
data). ``FPN w Finetune'' is the FPN model first trained on VG,
then fintunedg on MSCOCO with less data. The Universal-RCNN is better
than baseline methods with a large margin in all scenarios. Moreover,
it can be found that Universal-RCNN trained on half of the data (42\%
mAP) outperforms FPN trained on full data (39.2\% mAP). This superior
performance confirms the effectiveness of our method that seamlessly
bridges the gap between domains and fully utilizes data annotations.
Thus, our method can be quickly adapted to new dataset with less annotation
by borrowing information from other domains.

\begin{table}
\begin{centering}
\tabcolsep 0.01in{\small{}}%
\begin{tabular}{c|c|ccc}
\hline 
{\scriptsize{}\%} & {\scriptsize{}Method} & {\scriptsize{}AP} & {\scriptsize{}AP$_{50}$} & {\scriptsize{}AP$_{75}$}\tabularnewline
\hline 
\multirow{4}{*}{{\scriptsize{}\begin{sideways}{\scriptsize{}MSCOCO}\end{sideways}}} & {\scriptsize{}SIN  \parencite{liu2018structure}} & {\scriptsize{}23.2} & {\scriptsize{}44.5} & {\scriptsize{}22.0}\tabularnewline
 & {\scriptsize{}Relation Network  \parencite{Hu2018}} & {\scriptsize{}38.9} & {\scriptsize{}60.5} & {\scriptsize{}43.3}\tabularnewline
 & {\scriptsize{}HKRM  \parencite{jiang2018hybrid}} & {\scriptsize{}37.8} & {\scriptsize{}58.0} & {\scriptsize{}41.3}\tabularnewline
 & {\scriptsize{}Ours w Intra-Domain} & \textbf{\scriptsize{}41.4} & \textbf{\scriptsize{}60.9} & \textbf{\scriptsize{}46.0}\tabularnewline
\hline 
\multirow{2}{*}{{\scriptsize{}\begin{sideways}{\scriptsize{}VG}\end{sideways}}} & {\scriptsize{}HKRM  \parencite{jiang2018hybrid}} & {\scriptsize{}7.8} & {\scriptsize{}13.4} & {\scriptsize{}8.1}\tabularnewline
 & {\scriptsize{}Ours w Intra-Domain} & \textbf{\scriptsize{}7.9} & \textbf{\scriptsize{}13.7} & \textbf{\scriptsize{}8.3}\tabularnewline
\hline 
\multirow{2}{*}{{\scriptsize{}\begin{sideways}{\scriptsize{}ADE}\end{sideways}}} & {\scriptsize{}HKRM  \parencite{jiang2018hybrid}} & {\scriptsize{}10.3} & {\scriptsize{}18.0} & {\scriptsize{}10.4}\tabularnewline
 & {\scriptsize{}Ours w Intra-Domain} & \textbf{\scriptsize{}14.0} & \textbf{\scriptsize{}23.1} & \textbf{\scriptsize{}14.9}\tabularnewline
\hline 
\end{tabular}{\small\par}
\par\end{centering}
\caption{\label{tab:Intra-Reasoning Results}Comparison between intra-domain
semantic correlation methods. The backbone are ResNet-101.}
\end{table}

\textbf{Comparison with semantic correlation works within one domain.}
There have been many works that use graph based models for additional
reasoning over bounding boxes, we compare our intra-domain graph reasoning
method with those intra-domain semantic correlation works in this
section. We train FPN with our intra-domain module using one dataset
and report the results in Table \ref{tab:Intra-Reasoning Results}.
It can be found that our Intra-Domain graph reasoning module is superior
to multiple competing methods. It should be noted that previous works \parencite{hu2017relation,jiang2018hybrid,liu2018structure}
use fully connected graphs to build object-object relationships, our
method instead learn a sparse spatial-aware graph structure to perform
graph inference, which can reduce lots of redundant relationships
and improve the detection performance significantly. Furthermore,
compare to the implicit knowledge module in  \citeauthor{jiang2018hybrid},
our method uses global semantic pool to construct bbox features and
use a GCN to aggregate information. 

\begin{table}
\begin{centering}
\tabcolsep 0.02in{\small{}}%
\begin{tabular}{c|c|ccccccc}
\hline 
\multirow{2}{*}{{\scriptsize{}\%}} & \multirow{2}{*}{{\scriptsize{}Method}} & \multirow{2}{*}{{\scriptsize{}backbone}} & \multicolumn{6}{c}{\textit{\scriptsize{}test-val}}\tabularnewline
\cline{4-9} \cline{5-9} \cline{6-9} \cline{7-9} \cline{8-9} \cline{9-9} 
 &  &  & {\scriptsize{}AP} & {\scriptsize{}AP$_{50}$} & {\scriptsize{}AP$_{75}$} & {\scriptsize{}AP$_{S}$} & {\scriptsize{}AP$_{M}$} & {\scriptsize{}AP$_{L}$}\tabularnewline
\hline 
\multirow{13}{*}{{\scriptsize{}\begin{sideways}{\scriptsize{}MSCOCO}\end{sideways}}} & {\scriptsize{}DetNet} & {\scriptsize{}DetNet-59} & {\scriptsize{}35.1} & {\scriptsize{}57.3} & {\scriptsize{}37.2} & {\scriptsize{}22.3} & {\scriptsize{}40.8} & {\scriptsize{}39.9}\tabularnewline
 & {\scriptsize{}DetNet w MB} & {\scriptsize{}DetNet-59} & {\scriptsize{}36.8} & {\scriptsize{}59.0} & {\scriptsize{}39.3} & {\scriptsize{}22.6} & {\scriptsize{}42.3} & {\scriptsize{}42.3}\tabularnewline
 & {\scriptsize{}DetNet w ours} & {\scriptsize{}DetNet-59} & \textbf{\scriptsize{}40.1} & \textbf{\scriptsize{}61.8} & \textbf{\scriptsize{}43.2} & \textbf{\scriptsize{}24.7} & \textbf{\scriptsize{}46.5} & \textbf{\scriptsize{}46.5}\tabularnewline
\cline{2-9} \cline{3-9} \cline{4-9} \cline{5-9} \cline{6-9} \cline{7-9} \cline{8-9} \cline{9-9} 
 & {\scriptsize{}HKRM} & {\scriptsize{}ResNet-101} & {\scriptsize{}40.1} & {\scriptsize{}61.2} & {\scriptsize{}44.8} & {\scriptsize{}23.8} & {\scriptsize{}43.7} & {\scriptsize{}51.2}\tabularnewline
 & {\scriptsize{}HKRM w MB} & {\scriptsize{}ResNet-101} & {\scriptsize{}41.6} & {\scriptsize{}62.9} & {\scriptsize{}45.5} & {\scriptsize{}24.3} & {\scriptsize{}45.2} & {\scriptsize{}52.8}\tabularnewline
 & {\scriptsize{}HKRM w ours} & {\scriptsize{}ResNet-101} & \textbf{\scriptsize{}43.7} & \textbf{\scriptsize{}63.5} & \textbf{\scriptsize{}47.3} & \textbf{\scriptsize{}24.8} & \textbf{\scriptsize{}46.5} & \textbf{\scriptsize{}55.4}\tabularnewline
\cline{2-9} \cline{3-9} \cline{4-9} \cline{5-9} \cline{6-9} \cline{7-9} \cline{8-9} \cline{9-9} 
 & {\scriptsize{}Mask-RCNN} & {\scriptsize{}ResNet-101} & {\scriptsize{}39.7} & {\scriptsize{}61.6} & {\scriptsize{}43.3} & {\scriptsize{}23.1} & {\scriptsize{}43.3} & {\scriptsize{}49.7}\tabularnewline
 & {\scriptsize{}Mask-RCNN w MB} & {\scriptsize{}ResNet-101} & {\scriptsize{}41.0} & {\scriptsize{}62.2} & {\scriptsize{}45.0} & {\scriptsize{}24.2} & {\scriptsize{}44.9} & {\scriptsize{}51.0}\tabularnewline
 & {\scriptsize{}Mask-RCNN w ours} & {\scriptsize{}ResNet-101} & \textbf{\scriptsize{}45.1} & \textbf{\scriptsize{}65.2} & \textbf{\scriptsize{}49.3} & \textbf{\scriptsize{}26.1} & \textbf{\scriptsize{}48.6} & \textbf{\scriptsize{}57.3}\tabularnewline
\cline{2-9} \cline{3-9} \cline{4-9} \cline{5-9} \cline{6-9} \cline{7-9} \cline{8-9} \cline{9-9} 
 & {\scriptsize{}FPN} & {\scriptsize{}ResNet-101} & {\scriptsize{}38.9} & {\scriptsize{}61.0} & {\scriptsize{}42.3} & {\scriptsize{}22.3} & {\scriptsize{}42.3} & {\scriptsize{}48.5}\tabularnewline
 & {\scriptsize{}FPN w MB} & {\scriptsize{}ResNet-101} & {\scriptsize{}40.1} & {\scriptsize{}61.7} & {\scriptsize{}43.6} & {\scriptsize{}23.2} & {\scriptsize{}43.8} & {\scriptsize{}49.7}\tabularnewline
 & {\scriptsize{}FPN w ours} & {\scriptsize{}ResNet-101} & \textbf{\scriptsize{}44.4} & \textbf{\scriptsize{}64.9} & \textbf{\scriptsize{}48.5} & \textbf{\scriptsize{}25.3} & \textbf{\scriptsize{}47.8} & \textbf{\scriptsize{}56.7}\tabularnewline
 & {\scriptsize{}FPN w ours {*}} & {\scriptsize{}ResNext-101} & \textbf{\scriptsize{}49.1} & \textbf{\scriptsize{}69.3} & \textbf{\scriptsize{}54.2} & \textbf{\scriptsize{}30.4} & \textbf{\scriptsize{}52.7} & \textbf{\scriptsize{}61.7}\tabularnewline
\hline 
\end{tabular}{\small\par}
\par\end{centering}
\begin{centering}
{\small{}}{\small\par}
\par\end{centering}
\caption{\label{tab:mAP-COCO}Methods trained on multiple datasets(COCO, VG,
ADE) and eval on COCO. ``MB'' is model with Multi Branches. {*}
is the model adding multi-scale training and testing with Soft-NMS}
\end{table}

\textbf{Generalization Capacity. }To validate the generalization capability
of the Universal-RCNN, we further implement our method upon more recent
detection methods such as DetNet \parencite{li2018detnet}, HKRM  \parencite{jiang2018hybrid}
amd Mask-RCNN  \parencite{he2017mask} and compair. The results are
reported on COCO in Table \ref{tab:mAP-COCO}. Table \ref{tab:mAP-COCO}
shows that Universal-RCNN consistently boost the performance by 4\textasciitilde 6
points in terms of AP\textit{, }which\textit{ }suggestes that the\textit{
}Universal-RCNN is widely applicable across different detection baselines.
We futher add some bells and whistles to test the upper limit of the
Universal-RCNN  \parencite{li2018detnet}. Specifically, we utilize
ResNext-101 as the backbone and apply multi-scale training, multi-scale
testing and Soft-NMS. Finally, the Universal-RCNN obtains 49.1 mAP
on COCO \textit{test-dev} with single-model result.

\section{Conclusion}

In this work, we proposed a new universal object detector (Universal-RCNN)
to alleviate the categories discrepancy and fully utilize the data
annotation. Our method can be easily plugged into any existing detection
pipeline via Transferable Graph R-CNN with multiple domains for endowing
its ability to global reasoning and transferring. Extensive experiments
demonstrate the effectiveness of the proposed method and achieve the
state-of-the-art results on multiple object detection benchmarks. 

\fontsize{9.0pt}{10.0pt} \selectfont
\bibliographystyle{aaai} \bibliography{1266}
 
\end{document}